\documentclass[11pt,twocolumn]{article}
\usepackage[]{fullpage}

\usepackage[utf8]{inputenc}
\usepackage[T1]{fontenc}
\usepackage[dvipsnames,table]{xcolor}
\usepackage[colorlinks=true,citecolor=Blue,urlcolor=Blue,linkcolor=Blue]{hyperref}
\usepackage{url}
\usepackage{booktabs}
\usepackage{amsfonts}
\usepackage{nicefrac}
\usepackage{microtype}
\usepackage{booktabs}
\usepackage{multirow}
\usepackage[export]{adjustbox}
\usepackage[caption=false]{subfig}
\usepackage{capt-of}
\usepackage{amsmath}
\usepackage{mathtools}
\usepackage{Definitions}
\usepackage{makecell}
\usepackage{fixltx2e}
\usepackage{enumitem}
\usepackage{color}
\usepackage[normalem]{ulem}
\usepackage{graphicx}

\usepackage[caption=false]{subfig}
\usepackage{capt-of}

\usepackage{authblk}

\begin{document}

\title{On Fairness, Diversity and \\Randomness in Algorithmic Decision Making}

\date{}
\makeatletter
\renewcommand\AB@affilsepx{, \protect\Affilfont}
\makeatother

\author[1]{Nina Grgi\'{c}-Hla\v{c}a}
\author[1]{Muhammad Bilal Zafar}
\author[1]{Krishna P. Gummadi}
\author[2,3,4]{Adrian Weller}
\affil[1]{\small Max Planck Institute for Software Systems (MPI-SWS)}
\affil[2]{\small University of Cambridge}
\affil[3]{\small Alan Turing Institute}
\affil[4]{\small Leverhulme Centre for the Future of Intelligence}

\maketitle

\begin{abstract}

Consider a binary decision making process where a single machine learning
classifier replaces a multitude of humans.
We raise questions about the
resulting loss of diversity in the decision making process.
We study the potential benefits of using random classifier ensembles instead of a single classifier
in the context of fairness-aware learning and demonstrate various attractive  properties:
(i) an ensemble of fair classifiers is guaranteed to be fair, for several different measures of fairness, (ii) an ensemble of unfair classifiers can still achieve fair outcomes, and (iii) an ensemble of classifiers can achieve better accuracy-fairness trade-offs than a single classifier.
Finally, we introduce notions of distributional fairness
to characterize further potential benefits of
random classifier ensembles.
\end{abstract}

\section {Introduction} \label{sec:intro}

A number of recent works have examined fairness concerns arising from
the recent trend of replacing human decision makers with systems based on machine learning
in scenarios ranging from
recidivism risk estimation~\cite{propublica_story,dimpact_fpr,zafar_dmt} and welfare benefit
eligibility~\cite{polish_unemployment} to loan approvals and credit
scoring~\cite{hardt_nips16}. However, these studies have largely overlooked the
{\it implicit loss in decision process diversity} that results from
replacing a large number of human decision makers, each of whom might have their own distinct decision criteria,
with a single
decision making algorithm.

When humans make decisions, the decision process diversity is inevitable due to our limited cognitive capacities.
For instance, no single human
judge can possibly estimate recidivism risk for all criminals in a city or
country. Consequently, in practice,
individual cases are assigned to a {\it
  randomly selected} sub-panel of one or more {\it randomly selected}
judges~\cite{judgeassignment_mass,judgeassignment_mn}. Random assignment
is key to
achieving fair treatment,
as different sub-panels of human judges might make decisions differently
and each case should have an equal chance of being judged by every
possible sub-panel.

In contrast,
a single decision making algorithm can
be scaled easily to handle any amount of workload by simply adding
more computing resources.
Current practice is to
replace a multitude of human decision makers with a single algorithm, such as
COMPAS for recidivism risk estimation in the U.S.~\cite{propublica_story} or
the algorithm introduced by the Polish Ministry of Labor and Social Policy, used for welfare benefit eligibility decisions in Poland~\cite{polish_unemployment}.
However, we remark that one could  introduce
diversity into machine decision making by instead training a collection of
algorithms (each might capture a different
``school of thought''~\cite{welinder2010multidimensional} as used by judges),
randomly assigning a case to a subset, then combining
their decisions in some ensemble manner (\eg, simple or weighted majority
voting, or unanimous consensus). Another motivation for exploring such
approaches is the rich literature on ensemble learning,
where a combination of a diverse ensemble of predictors may been shown (both
theoretically and empirically) to outperform single predictors
on a variety of tasks~\cite{brown2005diversity}.

Against this background, we explore the following question: {\it for
  the purposes of fair decision making, are there any fundamental
  benefits to replacing a single decision making algorithm with a
  diverse ensemble of decision making algorithms?} In this paper, we
consider the question in a restricted set of scenarios, where the
algorithm is a binary classifier and the decisions for any given user
are made by a single randomly selected classifier from the
ensemble.
While restrictive, these scenarios capture decision making
in a number of real-world settings (such as a randomly assigned judge
deciding whether or not grant bail to an applicant)
and reveal striking results.

Our findings, while preliminary, show that compared to a single
classifier, a diverse ensemble can not only achieve better fairness in
terms of distributing beneficial outcomes more uniformly amongst the
set of deserving users, but can also achieve better accuracy-fairness
trade-offs for existing notions (measures) of unfairness such as
disparate treatment \cite{salvatore_knn,zafar_fairness, icml2013_zemel13}, impact \cite{feldman_kdd15, zafar_fairness, icml2013_zemel13}, and mistreatment \cite{hardt_nips16, zafar_dmt}.
Interestingly,
we find that for certain notions of fairness, a diverse ensemble is
not guaranteed to be fair even when individual classifiers within the
ensemble are fair.
On the other hand, a diverse ensemble can be fair
even when the individual classifiers comprising the ensemble are
unfair. Perhaps surprisingly, we show that it is this latter property
which enables a diverse ensemble of individually unfair classifiers to
achieve better accuracy-fairness trade-offs than any single classifier.

Our work suggests that further research in the area of ensemble-based methods may be very fruitful when
designing
fair learning mechanisms.

\section {Fairness of classifier ensembles} \label{sec:sec1}

We first introduce our ensemble approach (randomly selecting a classifier from a diverse set) and various notions of fairness in
classification, then demonstrate interesting, and perhaps surprising, properties of the ensemble classifiers.
We assume exactly one sensitive attribute (\eg, gender or race) which is binary, though the results may naturally be extended beyond a single binary attribute.

Assume we have an ensemble $\Ccal_{\text{ens}}$
of $M$ individual
classifiers $\{\Ccal_j\}_{j=1}^{M}$, operating on a dataset $\{(\xb_i,
y_i, z_i)\}_{i=1}^{N}$.
Here, $\xb_i \in \RR^d$, $y_i \in \{-1, 1\}$
and $z_i \in \{0, 1\}$ respectively denote the feature vector, class
label and sensitive attribute value of the $i^{th}$ user.
Each classifier
$\Ccal_j: \RR^d \to \{-1, 1\}$ maps a
given user feature vector $\xb_i$ to a predicted outcome
$\hat{y}_{i,j}$, \ie, $\hat{y}_{i,j} = \Ccal_j(\xb_i)$.
We assume we are given a probability distribution $p(j)$ over the classifiers. Overloading notation, we consider the ensemble classifier $\Ccal_{\text{ens}}$ defined to operate on $\xb_i$ by first  selecting some $\Ccal_{k} \in \Ccal_{\text{ens}}$ independently at random according to the distribution $p$, and then returning $\Ccal_{\text{ens}}(\xb_i) = \Ccal_{k}(\xb_i) \in \{-1,1\}$.
\footnote{Equivalently, this may be considered an ensemble approach where  $\hat{y}_{i,j}=\Ccal_j(\xb_i)$ is computed $\forall j$, then we randomly output 1 or $-1$ with respective probabilities $\sum_{j:\hat{y}_{i,j}=1} p(j)$ or $\sum_{j:\hat{y}_{i,j}=-1} p(j)$.}

Two common notions used to assess the fairness of a decision making system require that a classifier should provide~\cite{barocas_2016}:
(1) {\it Equality of treatment}, \ie, its prediction for a user should not depend on the user's sensitive attribute value (\eg, man, woman); and/or (2) {\it Equality of impact}, \ie, rates of beneficial outcomes should be the same for all sensitive attribute value groups (\eg, men, women).
For (2), various
measures of beneficial outcome rates have been proposed:
acceptance rates into the positive (or negative) class for the group \cite{feldman_kdd15, zafar_fairness, icml2013_zemel13}; the classifier’s true positive (or negative) rate for the group \cite{hardt_nips16, klein16, zafar_dmt}; or the classifier’s predictive positive (or negative) rate—also called positive (or negative) predictive value—for the group~\cite{klein16,zafar_dmt}.
For a discussion on these measures, see~\cite{klein16,zafar_dmt}.

\subsection{Is an ensemble of fair classifiers guaranteed to be fair? In many cases, yes.}
For any ensemble $\Ccal_{\text{ens}}$ consisting of classifiers $\{\Ccal_j\}_{j=1}^{M}$ as above, it is immediate to see that if all $\Ccal_j$ satisfy equality of treatment, then $\Ccal_{\text{ens}}$ also satisfies equality of treatment.

Next, one can easily show that if all $\Ccal_j$ satisfy equality of impact (\ie, equality of beneficial outcome rates), where the beneficial outcome rates are defined as the acceptance rate into the positive (negative) class, or the true positive (negative) rate, then $\Ccal_{\text{ens}}$ will also satisfy the equality of impact.  For example, if beneficial outcome rates are defined in terms of  acceptance rate into the positive class, and
expected benefits are the same for all $\Ccal_j$:
\begin{align*}
\underset{\xb|z=0}{\EE}[\II\{\Ccal_j(\xb)\ = 1\}] = \underset{\xb|z=1}{\EE}[\II\{\Ccal_j(\xb)\ = 1\}] \quad \forall j,
\end{align*}
where $\II$ is the indicator function, then one can show that:
\begin{align*}
\underset{\xb|z=0}{\EE}[\II\{\Ccal_{\text{ens}}(\xb)\ = 1\}] = \underset{\xb|z=1}{\EE}[\II\{\Ccal_{\text{ens}}(\xb)\ = 1\}],
\end{align*}
using linearity of expectation since all expectations are defined over groups of constant sizes (left hand side defined over group with $z=0$ and right hand side over $z=1$). The same can be shown when beneficial outcome rates are defined in terms of true positive (negative) rates. That is, for the true positive rate,  if it holds that:
\begin{align*}
\underset{\xb|z=0,y=1}{\EE}[\II\{\Ccal_j(\xb)\ = 1\}] = \\
\underset{\xb|z=1,y=1}{\EE}[\II\{\Ccal_j(\xb)\ = 1\}] \quad \forall j,
\end{align*}
one can show that:
\begin{align*}
\underset{\xb|z=0,y=1}{\EE}[\II\{\Ccal_{\text{ens}}(\xb)\ = 1\}] = \\
\underset{\xb|z=1,y=1}{\EE}[\II\{\Ccal_{\text{ens}}(\xb)\ = 1\}].
\end{align*}

On the other hand,
this no longer holds if beneficial outcome rates are defined in terms of positive (negative) predictive value, since these values are computed as expectations over the size of the predicted positive or negative class of a classifier $\Ccal_j$. Specifically, the expected positive predictive value of a classifier $\Ccal_j$ for group with $z=0$ is defined as:
\begin{align*}
\underset{\xb,y|z=0, \Ccal_j(\xb) = 1}{\EE}[\II\{y=1\}].
\end{align*}
Since the expectation is defined over $\Ccal_j(\xb) = 1$, which changes for every $j \in [0,M]$, we can no longer apply linearity of expectation, and hence $\Ccal_{\text{ens}}$ will not in general satisfy this notion of equality of impact even when all $\Ccal_j$ do have this property.

\subsection{Can an ensemble of unfair classifiers be fair? Yes.}
For all fairness notions of equality of treatment or equality of impact described above, there exist cases where a random ensemble of unfair classifiers can indeed be fair.
Here we show examples of such cases for equality of treatment, and for equality of impact (or the equality of beneficial outcome rate) where the benefit outcome rate is defined as the fraction of users from a sensitive attribute value groups (\eg, men, women) accepted into the positive class~\cite{feldman_kdd15,zafar_fairness,icml2013_zemel13}. Examples where the benefit measure is defined in terms or error rates~\cite{hardt_nips16,zafar_dmt} can be similarly constructed.

\begin{figure}[t]
\centering
\includegraphics[width=0.6\columnwidth]{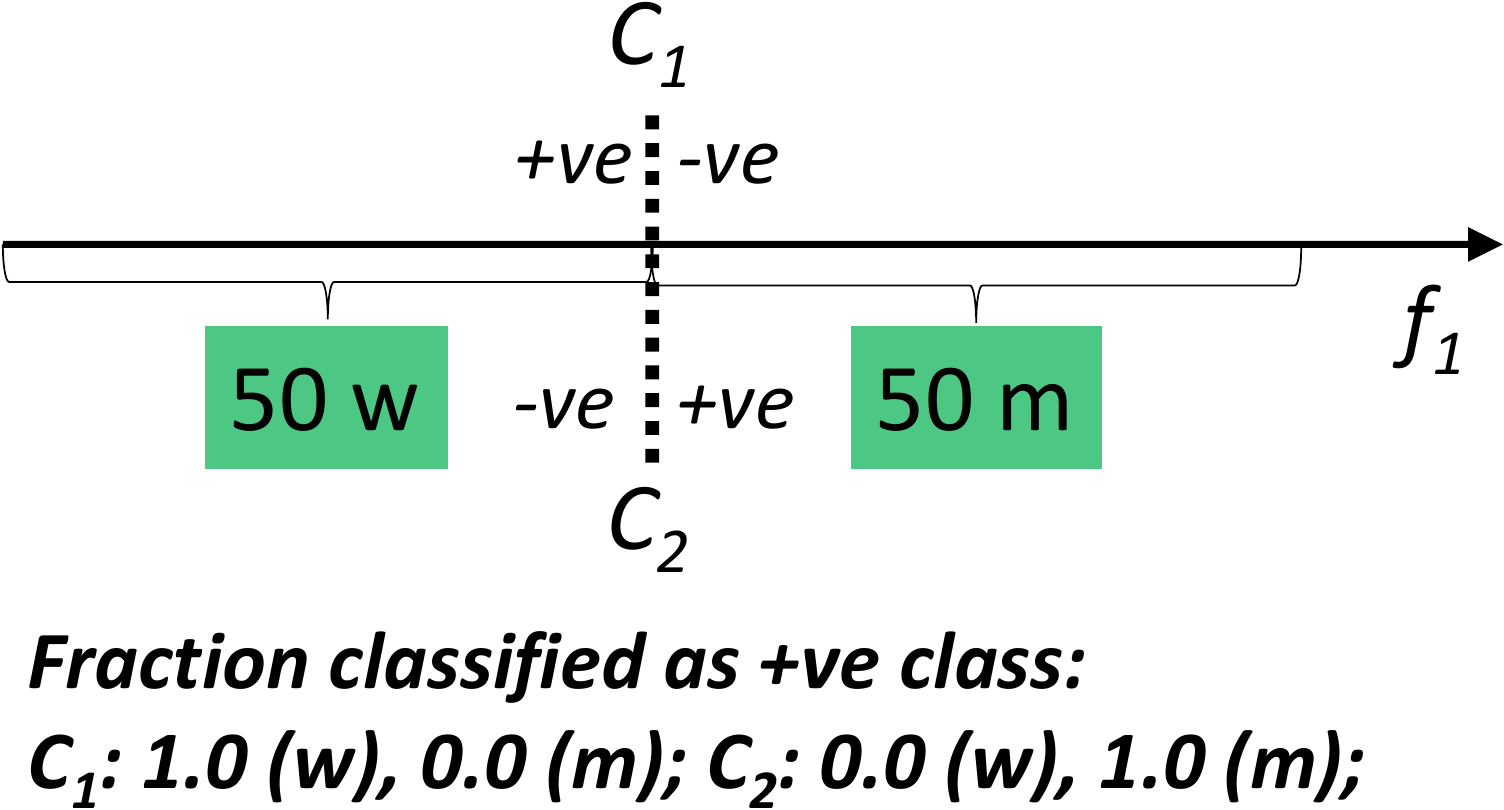}
\caption{
A fictitious decision making scenario involving two groups of people: men (m) and women (w); a single feature: $f_1$, which is gender in this case; and two classifiers: $\Ccal_1$ and $\Ccal_2$.
The classifiers $\Ccal_1$ and $\Ccal_2$
do not satisfy equality of treatment because their outcomes solely depend on the user's sensitive attribute value, \ie, $\Ccal_1$ ($\Ccal_2$) classifies all women (all men) into the positive class while classifying all men (all women) into the negative class.
On the other hand, an ensemble of these classifiers that
chooses classifier $\Ccal_1$ and $\Ccal_2$ with probability $\frac{1}{2}$ each
is fair because its decisions would not change based on the users' gender.
}
\label{pic4}
\end{figure}

\paragraph{Equality of treatment.}  Consider the example shown in Figure~\ref{pic4} which shows a decision making scenario involving two sensitive attribute value groups, men and women, and two classifiers $\Ccal_1$ and $\Ccal_2$. The equality in treatment fairness criterion requires that a
classifier must treat individuals
equally regardless of their sensitive attribute value (\ie, regardless of whether the subject being classified is a man or a woman).
Observe that neither $\Ccal_1$ nor $\Ccal_2$ satisfies
this criterion, since each accepts only women or men, respectively. On the other hand, an ensemble of these classifiers that chooses $\Ccal_1$ and $\Ccal_2$ uniformly at random satisfies equality of treatment.

\begin{figure}[t]
\centering
\includegraphics[width=0.6\columnwidth]{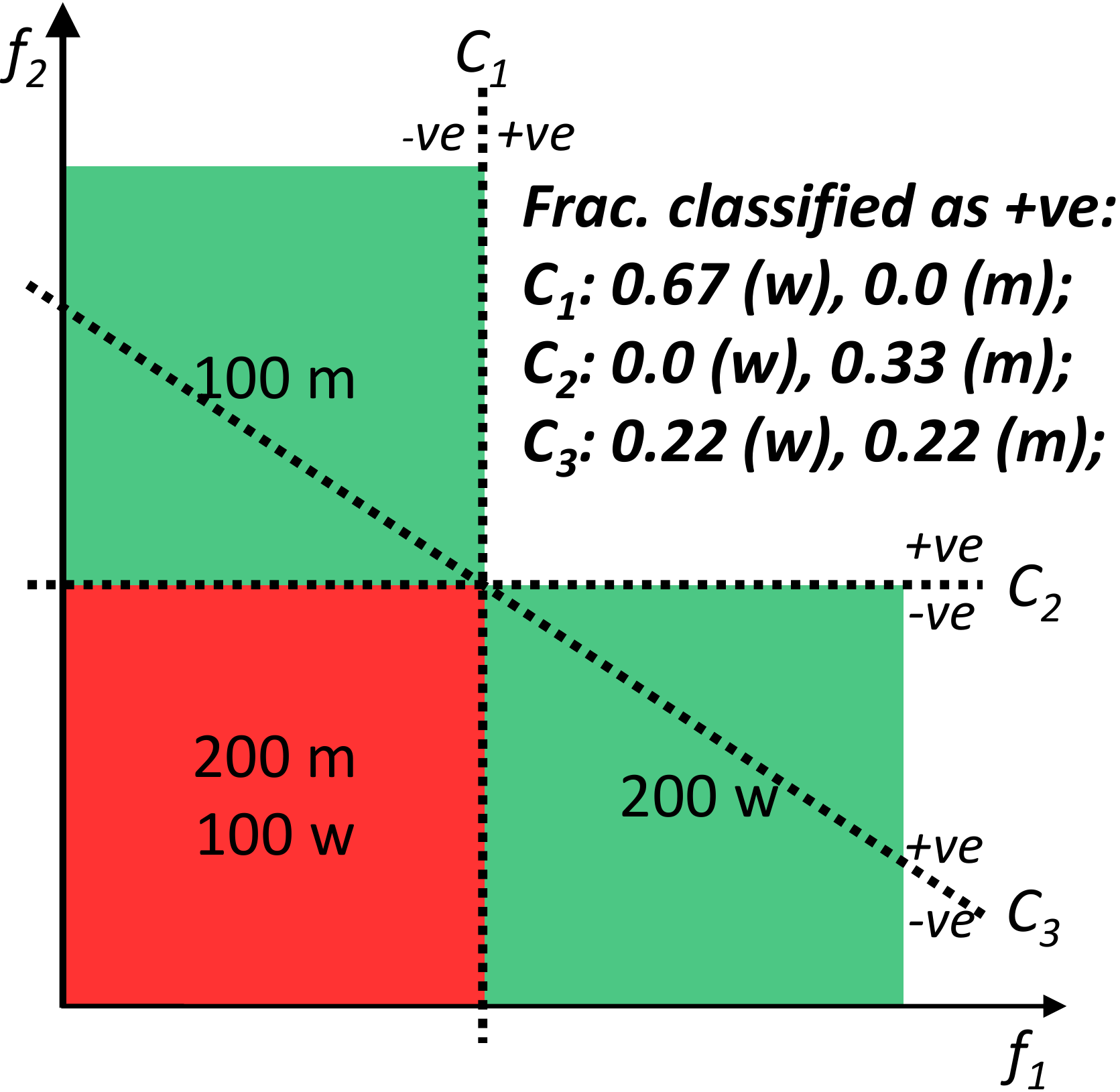}
\caption{
A  decision making scenario involving two groups of people: men (m) and women (w); two features: $f_1$ and $f_2$; and three classifiers: $\Ccal_1$, $\Ccal_2$ and $\Ccal_3$.
Green quadrants indicate the ground truth positive class in the training data, while red quadrants indicate the respective negative class.
Within each quadrant, the points are distributed uniformly. Gender is not one of the features ($f_1$ and $f_2$) used by the classifiers.
Classifiers $\Ccal_1$ and $\Ccal_2$
do not meet the equality of impact criterion (when group benefits are measured as rates of positive class acceptance) since they assign only men and only women to the positive class, respectively.
$\Ccal_3$ is a fair classifier by this measure,
since it gives both men and women the same $0.22$ positive class acceptance rate.
Let $\Ccal_{\text{ens}}$ be an ensemble  that
selects classifier $\Ccal_1$ with probability $\frac{1}{3}$, and  classifier $\Ccal_2$ with probability $\frac{2}{3}$. The ensemble, while consisting of  unfair classifiers, produces outcomes that are fair: it has the same $0.22$ positive class acceptance rate for both men and women.
}
\label{pic2}
\end{figure}

\paragraph{Equality of impact.}
We provide an example in Figure~\ref{pic2} where the impact fairness benefit measure is the rate of acceptance into the positive class.
Comparing the group benefits
given by $\Ccal_1$ and $\Ccal_2$,
both classifiers fail the
fairness criterion since they have different positive class acceptance rates for men and women (shown in the figure).
However, an ensemble
which  selects $\Ccal_1$  with probability $\frac{1}{3}$ and $\Ccal_2$
with probability $\frac{2}{3}$,
achieves the same acceptance rate for both women and men (since $\frac{1}{3} \times \frac{2}{3} \text{ women} + \frac{2}{3} \times 0 \text{ women} = \frac{2}{9} = \frac{1}{3} \times 0 \text{ men} + \frac{2}{3} \times \frac{1}{3} \text{ men}$).

\subsection{Can an ensemble of classifiers achieve better
  accuracy-fairness trade-offs than a single classifier? Yes.}
First, observe that by its definition, the accuracy of $\Ccal_{\text{ens}}$ is the expectation over the classifier probabilities $p(j)$ of the accuracies of the individual classifiers $\Ccal_{j}$.

When an individual classifier is optimized for accuracy subject to a fairness constraint, a significant loss of accuracy relative to the optimal unconstrained classifier may be unavoidable. If an ensemble is used instead, then we expand our model class to admit combinations of several unfair classifiers, some of which may have significantly higher accuracy than the optimal fair classifier
---requiring only that the ensemble classifier be fair.

\begin{figure}[t]
\centering
\includegraphics[width=0.9\columnwidth]{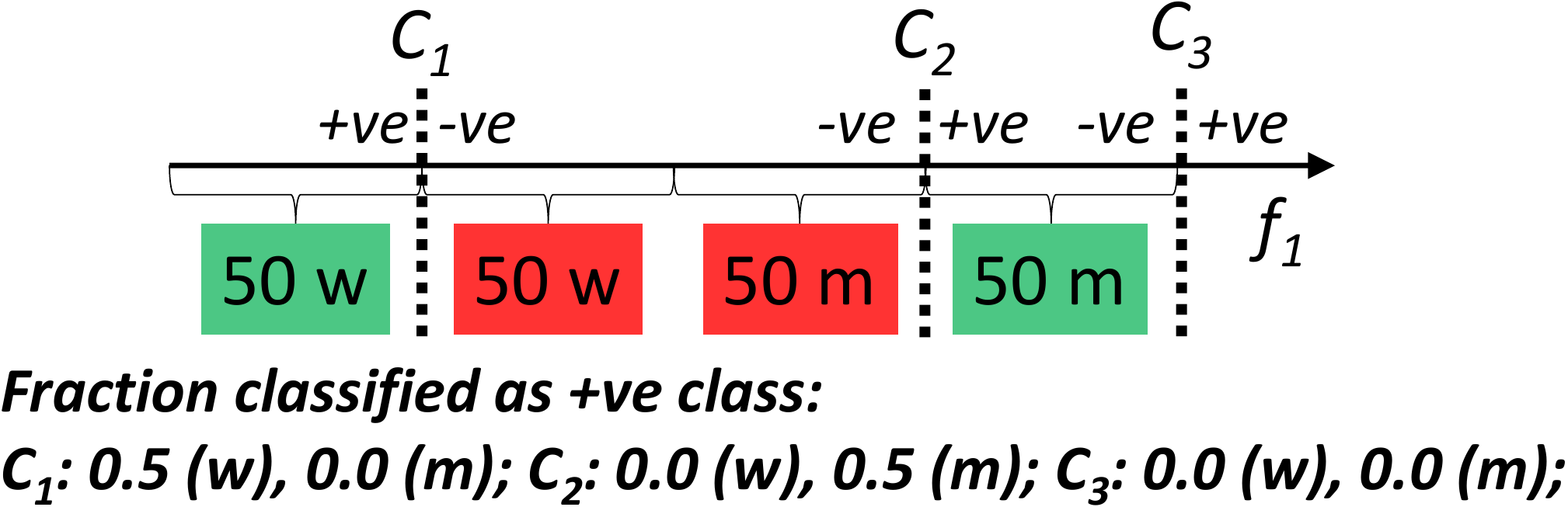}
\caption{
A decision making scenario involving one feature $f_1$, and three classifiers: $\Ccal_1$, $\Ccal_2$ and $\Ccal_3$.
A higher value of $f_1$ indicates the positive class (green) in the training data for men, but the negative (red) class for women.
In this scenario, no individual linear classifier can outperform $50\%$ accuracy, if we require equal benefits for both groups (where benefits are measured as rates of positive class acceptance).
However, an ensemble of $\Ccal_1$ and $\Ccal_2$ which selects each of them with $\frac{1}{2}$ probability, achieves fairness (equality in benefits)
with a much better accuracy of $0.75$.
}
\label{pic3}
\end{figure}

We provide an example in~Figure~\ref{pic3}.
We consider fairness as determined by equality of rates of positive class acceptance for men and women. Given the distribution of data shown, for a single classifier to be fair, it must be either at the extreme left (everyone is classified as positive) or at the extreme right (everyone is classified as negative)---in either case with accuracy of $50\%$, which in this example is the optimal achievable accuracy for a single fair classifier.

Now consider an ensemble of the two classifiers $\Ccal_1$ and $\Ccal_2$ shown, selecting either one with probability $\frac{1}{2}$.
This ensemble satisfies the fairness criterion (with positive rates of $0.25$ for each sex) and has an accuracy of $75\%$, which is much better than the single fair classifier optimum of $50\%$.

\subsection{Notions of \emph{distributional fairness}}
The behavior of an ensemble classifier differs from its
constituent classifiers in subtle but important ways. In particular,
for data points (or individual users)
on which
the constituent classifiers yield
different outcomes, our approach of randomly selecting a single
classifier introduces non-determinism in classifier output, \ie, there is a non-zero chance
of
both beneficial and
non-beneficial outcomes.

\begin{figure}[t]
\centering
\includegraphics[width=0.6\columnwidth]{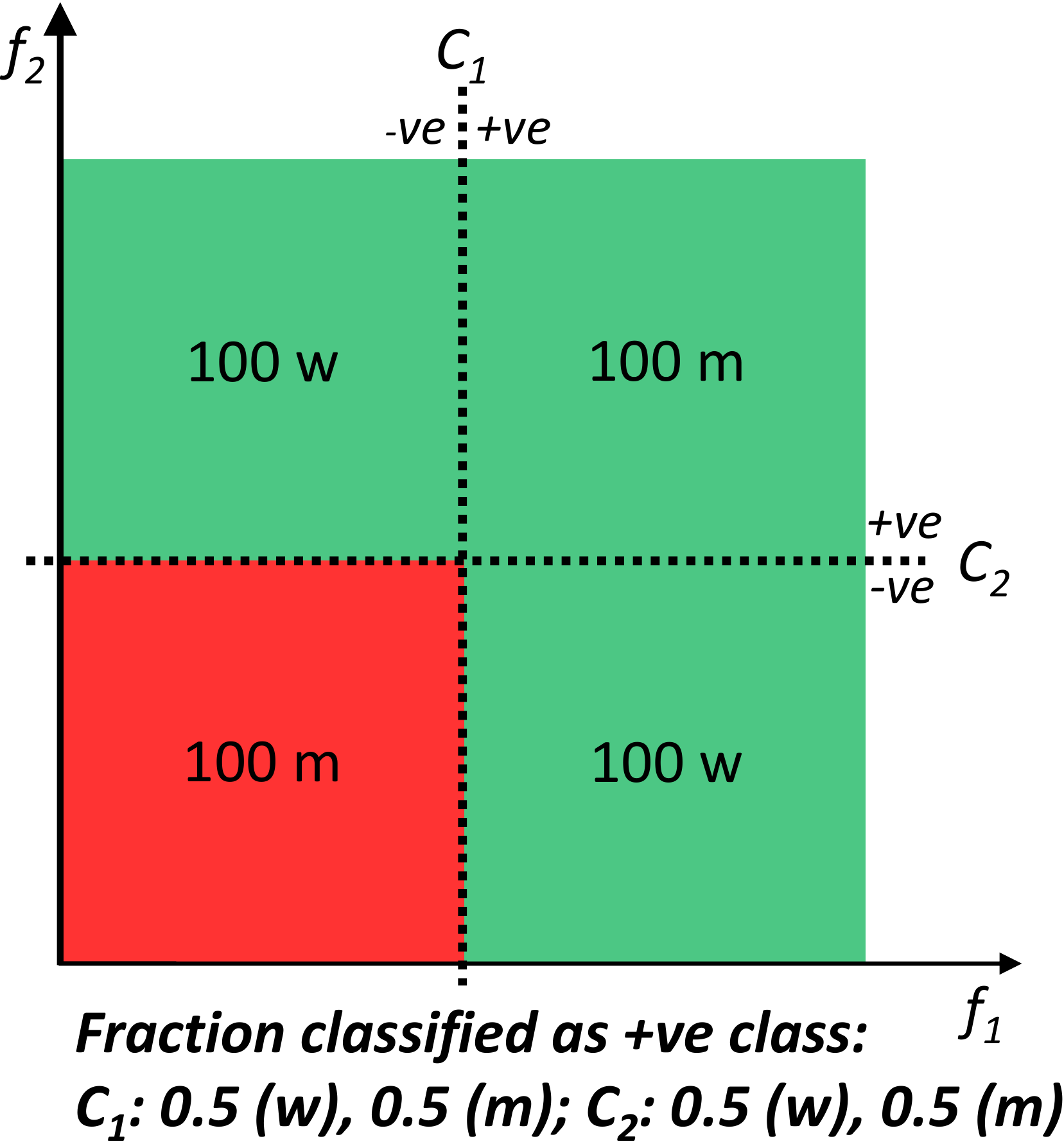}
\caption{
Classifiers $\Ccal_1$ and $\Ccal_2$ satisfy equality of impact, since their beneficial outcome rates (defined as the rates of positive class acceptance) are the same for men and women.
Consider an ensemble of the two classifiers which chooses each of $\Ccal_1$ and $\Ccal_2$ uniformly at random.
The ensemble also satisfies equality of impact, yet the \emph{distribution} of beneficial outcomes is very different among men and women:
half the men (top right quadrant) always get the positive outcome, while half the men (bottom left) always get the negative outcome;
whereas every woman gets the positive outcome randomly with probability 0.5.}
\label{pic1}
\end{figure}

We illustrate this scenario in
Figure~\ref{pic1}, showing
two classifiers $\Ccal_1$ and
$\Ccal_2$, where each has fair impact
in that both $\Ccal_1$ and
$\Ccal_2$ assign beneficial outcomes (positive class outcomes in this case) deterministically to 50\% of men and 50\% of women. However,
the classifiers differ in terms of the set of women that are assigned
the beneficial outcomes. By creating a $\frac{1}{2}:\frac{1}{2}$ ensemble of both classifiers,
we ensure instead
that \emph{all women} have an equal (50\%) chance of the
beneficial
outcome (while we still satisfy equality in impact).

This suggests an interesting question:
{\it
Are the outcomes of the ensemble classifier more fair than those of
the individual classifiers ($\Ccal_1$ and $\Ccal_2$) that comprise it?}

While all these classifiers satisfy the equality in impact fairness constraint,
one could make the case that the ensemble is more fair as it offers
all women an {\it equal} chance at getting beneficial outcomes,
whereas $\Ccal_1$ and $\Ccal_2$ pre-determine the subset of women who will
get the beneficial outcomes.

To our knowledge, no existing measure of algorithmic fairness captures this  notion of
evenly distributing beneficial outcomes across all
members of an
attribute group. Rather,
existing fairness
measures focus on fair assignment of outcomes between sensitive groups
({\it inter-group} fairness), while largely ignoring
fair assignment of outcomes within a sensitive group ({\it
intra-group} fairness).

These observations suggest the need in future work for new notions of distributional fairness to characterize the benefits achievable
with diverse classifier ensembles.

\section {Discussion} \label{sec:conclusion}

We have begun to explore the properties of using a random ensemble of classifiers in fair decision making, focusing on randomly selecting one classifier from a diverse set. It will be interesting in future work to explore a broader set of ensemble methods. Fish et al.~\cite{fish_boosting} examined fairness when constructing a deterministic classifier using boosting, but
we are not aware of prior work in fairness which considers how randomness in ensembles may be helpful.

We note a similarity to a Bayesian perspective: rather than aiming for the one true classifier, instead we work with a probability distribution over possible classifiers. An interesting question for future work is how to update the distribution over classifiers as more data becomes available, noting that we
may want to maintain diversity~\cite{GhaKim12}.

Decision making systems consisting of just one classifier facilitate the  ability of users to game the system. On the other hand, in an ensemble scheme such as the one we consider where a classifier is randomly selected: if an individual aims to achieve
some high threshold level of probability of a good classification, first she must acquire knowledge about the whole set of classifiers and the probability distribution over them, and then she must attain features some distance beyond the expected decision boundary of the ensemble (`a fence around the law').

A common notion of fairness is that individuals with similar features should obtain similar outcomes. However, a single deterministic classifier boundary causes individuals who are just on either side to obtain completely different outcomes. Using instead a distribution over boundaries leads to a smoother, more robust profile of expected outcomes, highlighting another useful property of ensembles in the context of fair classification.

\section*{Acknowledgements} \label{sec:acknowledgements}
AW acknowledges support by the Alan Turing Institute under EPSRC grant EP/N510129/1, and by the Leverhulme Trust via the CFI.

\bibliographystyle{abbrv}

\end{document}